\documentclass[10pt,twocolumn,letterpaper]{article}

\usepackage[pagenumbers]{wacv} 

\usepackage{graphicx}
\usepackage{amsmath}
\usepackage{amssymb}
\usepackage{booktabs}
\usepackage[table]{xcolor} 

\usepackage{pifont}
\newcommand{\cmark}{\ding{51}}%
\newcommand{\xmark}{\ding{55}}%

\usepackage[pagebackref,breaklinks,colorlinks]{hyperref}

\usepackage[capitalize]{cleveref}
\crefname{section}{Sec.}{Secs.}
\Crefname{section}{Section}{Sections}
\Crefname{table}{Table}{Tables}
\crefname{table}{Tab.}{Tabs.}

\def\wacvPaperID{1806} 
\def\confName{WACV}
\def\confYear{2024}

\begin{document}

\title{M$^{3}$3D: Learning 3D priors using Multi-Modal Masked Autoencoders for 2D image and video understanding}

\author{Muhammad Abdullah Jamal and Omid Mohareri \\
Intuitive Surgical Inc., Sunnyvale, CA}
\maketitle

\begin{abstract}
 We present a new pre-training strategy called M$^{3}$3D (\underline Multi-\underline Modal \underline Masked \underline 3D) built based on Multi-modal masked autoencoders that can leverage 3D priors and learned cross-modal representations in RGB-D data. We integrate two major self-supervised learning frameworks; Masked Image Modeling (MIM) and contrastive learning; aiming to effectively embed masked 3D priors and modality complementary features to enhance the correspondence between modalities. In contrast to recent approaches which are either focusing on specific downstream tasks or require multi-view correspondence, we show that our pre-training strategy is ubiquitous, enabling improved representation learning that can transfer into improved performance on various downstream tasks such as video action recognition, video action detection, 2D semantic segmentation and depth estimation. Experiments show that M$^{3}$3D outperforms the existing state-of-the-art approaches on ScanNet, NYUv2, UCF-101 and OR-AR, particularly with an improvement of +1.3\% mIoU against Mask3D on ScanNet semantic segmentation. We further evaluate our method on low-data regime and demonstrate its superior data efficiency compared to current state-of-the-art approaches.   
\end{abstract}

\section{Introduction}
\label{sec:intro}

Pre-training Vision Transformers~\cite{vit} (ViTs) using Masked Autoencoders (MAE) followed by fine-tuning gives rise to the state-of-the-art results for various computer vision tasks such as image classification~\cite{mae,simmm,beit}, video activity recognition~\cite{mae_vid,videomae,videomae2}, semantic segmentation~\cite{mask3d,gao2022convmae,li2022semmae} and 3D scene understanding~\cite{pimae,zhang2022point,pang2022masked}. Inspired by BERT~\cite{BERT}, MAE masks high number of patches in an input image or video clip and predicts such missing regions. It usually consists of an asymmetric encoder-decoder architecture, where the encoder objective is to encode the unmasked patches into latent representations while the masked patches are predicted by the decoder using these representations. However, MAEs are mostly focused on learning from single-modality data (images or video clips) and can't leverage other data modalities that are present in commonly used RGB-D and Time-of-Flight (ToF) cameras.
The depth and point cloud data available in such cameras can provide an opportunity to learn geometric priors and avoid view-dependent effects for efficient representation learning. Particularly, various 3D understanding approaches~\cite{chen20224dcontrast,xie2020pointcontrast,zhang2021selfsupervised,hou2021exploring} have been leveraging RGB-D datasets for high-level scene understanding and low-level point matching~\cite{banani2021bootstrap,zhang2023pcrcg} tasks using contrastive learning. However, very little work has been done in exploring 3D priors for 2D image understanding with limited focus on tasks like semantic segmentation and depth estimation. 

Our goal is to embed 3D priors into ViTs based backbones for various downstream tasks that include but not limited to video activity recognition, video action detection, semantic segmentation and depth estimation, to effectively learn geometric and structural priors as well as cross-modal features. Recently, Mask3D~\cite{mask3d} effectively learns masked 3D priors from a single-view RGB-D frame by only reconstructing the depth through masking out patches in RGB and depth modalities. However, the approach is limited to 2D image understanding tasks and relies on MAE pre-training for any cross-modal representation learning. Pri3D~\cite{pri3d_} explores to embed the 3D prior by leveraging multi-view RGB-D data to introduce geometric constraints in a contrastive learning scheme for ResNet backbones. However, it relies on camera registration between multiple views for each frame. Instead, we consider to empower ViTs with such 3D priors using MAEs and cross-modal training strategies such as constrastive learning and matching loss that transfer well to various video and image based downstream tasks. Cross-modal learning and Masked Image Modeling are complementary to each other. While the former would leverage the very useful visual-depth pair information to encode modality-invariant features, the latter forces its representation to encode the majority of the input information. This motivates us to integrate both the pre-training strategies to extract different discriminative features.


In this paper, we present M$^{3}$3D, a multi-modal masked autoencoder based approach which learns to embed 3D priors in a self-supervised manner by pre-training on RGB-D image pair or video clips and also integrates cross-modal learning. To train M$^{3}$3D, we randomly mask patches in both RGB and depth modalities and encode the unmasked patches by passing them through either modality-specific transformer encoders or modality-agnostic transformer encoder which is shared between both the modalities. The representations from the encoder are then appended with learned masked tokens to reconstruct masked regions in depth and RGB input. We equipped our approach with contrastive learning and RGB-D matching which are used to  enhance cross-modal correspondence during pre-training. The objective of contrastive learning is to align the visual scene and the corresponding depth data by pulling the feature associated with the RGB patch and its corresponding depth while pushing away the other depth patches in the corresponding depth map. Finally, to further enhance the correspondence, the matching loss predicts whether the RGB-depth pair is matched (positive) or not matched (negative) by applying a linear layer followed by softmax on the top of the encoder features to predict a two-class probability. Our experiments demonstrate the effectiveness of M$^{3}$3D on a variety of datasets for video and image understanding.  We pre-train on UCF-101~\cite{ucf101}, OR-AR~\cite{OR-AR} and fine-tune for video activity recognition, and video action detection respectively. We also pre-train the model with ScanNet~\cite{scannet} and fine-tune it for semantic segmentation. We also show the generalizibility of the model on NYUv2~\cite{nyuv2} data for depth estimation besides semantic segmentation.
In summary, the main contributions of our paper are:
\begin{itemize}
    \item We propose a new self-supervised pre-training approach called M$^{3}$3D which is equipped with masked autoencoders and cross-modal training for RGB-D representation learning. Our method can be applied to various video and image understanding tasks based on single view RGB-D image pair or video clips.
    \item Our approach learns to embed 3D priors without any camera pose registration, while enhancing the cross-modal correspondence between RGB and depth modalities using contrastive learning and matching loss.
    \item We demonstrate the efficacy of M$^{3}$3D for current ViT backbones on variety of video (UCF-101, OR-AR surgical dataset) and image understanding (ScanNet, NYUv2) datasets.
    \item We further evaluate our method on low-data regime and demonstrate its superior data efficiency compared to current state-of-the-art approaches. Several ablation studies have been conducted and presented here to prove the effectiveness of various aspects of M$^{3}$3D.
\end{itemize}
\section{Related Work}
\label{sec:review}
Our work proposes a self-supervised learning algorithm for general computer vision tasks. We will review some of the approaches that are closely related to our approach from different aspects - Masked Autoencoder based pre-training for vision transformers, Multi-modal learning and pre-training by learning 3D priors.

\paragraph{Masked Autoencoder based pre-training for transformers.} Self-supervised learning paradigm learns visual features based on some pre-text task using large-scale unlabeled dataset, before fine-tuning for multiple downstream tasks. Earlier work on pre-text task includes colorizaton~\cite{zhang2016colorful}, jigsaw puzzle~\cite{jigsaw}, and predicting image rotation~\cite{rotnet}. Instance-discrimination based learning uses two different views or augmentations of an image to learn view-invariant features by pulling them closer in the embedding space while pushing away the negative pair. SimCLR~\cite{simclr}, MoCo~\cite{mocov2} uses contrastive learning, BYOL~\cite{byol} uses online and target network while SwAV~\cite{swav}, DeepCluster~\cite{deepcluster} uses clustering to discriminate between the clusters. Recently, the momentum in self-supervised learning has been shifted towards the masked image modeling which learns useful representations by learning to reconstruct the masked image patches. Inspired by the success in NLP domain such as bidirectional encoder (BERT)~\cite{BERT} and Generative Pre-Training (GPT)~\cite{GPT}, several masked image prediction methods for pre-training vision transformers have been proposed with various reconstruction targets such as pixels~\cite{SIT_method,vit,iGPT,elnouby2021largescale,mae,simmm}, features~\cite{maskfeat,data2vec}, discrete tokens via dVAE~\cite{beit,ibot}. They have shown to outperform constrastive learning based approaches on various downstream tasks. One particular approach is MAE~\cite{mae} which first masks large portion of the input (e.g., 75\%) and then passes the unmasked patches to the large encoder followed by a lightweight decoder that reconstruct the masked patches. By masking out large portion of input patches, it accelerates the pre-training of vision transformers. Moreover, these approaches have been extended to the video domain where most of focus is to design a masking strategy that includes random masking~\cite{mae_vid}, tube masking~\cite{videomae} and adaptive masking based on motion~\cite{surgmae, mme} or a separate network~\cite{adamae}. On the other-hand, our work proposes multi-modal masked auto-encoder pre-training by learning 3D priors for several downstream tasks such as video action detection, video activity recognition, 2D semantic segmentation and depth estimation.

\paragraph{Multi-Modal Learning.} It involves training the models by leveraging data from multiple modalities. The learning may involve training modality-agnostic unified encoder or modality-specific encoders using modalties such as images and text~\cite{alayrac2020selfsupervised,kaiser2017model,e2evlp,castrejon2016learning,hu2021unit,chen2020uniter,lxmert}, video and audio~\cite{arandjelović2017look, jaegle2022perceiver,nagrani2022attention,owens2018audiovisual}, video, text and audio~\cite{akbari2021vatt}, and images and depth~\cite{girdhar2022omnivore}. One particular approach is MultiMAE~\cite{multimae} which extends MAE to multi-modal multi-task settings and includes depth modality. However, it requires depth during the fine-tuning stage as well as semantic segmentation labels during the pre-training. Our approach doesn't rely on semantic segmentation, it only uses depth during the pre-training and it is not limited to the image domain only. Please refer to experimental section~\ref{sec:exps} for more results on MultiMAE without semantic segmentation modality.

\paragraph{Pre-training by learning 3D priors.} There is a recent surge of learning cross-modal features particularly between languages and images. CLIP~\cite{clip_} learns visual concepts from natural language supervision, showing impressive zero-shot performance for image classification. Pri3D~\cite{pri3d_} learns view-invariant and geometry-aware representations using multi-view consistency during pre-training for 2D image understanding. It leverages constrastive learning for 2D-3D correspondence and embed 3D priors to pre-train ResNet~\cite{ResNet} architectures. However, our approach is more ubiquitous, it is not limited to 2D image understanding tasks and doesn't require camera pose registration across views. Mask3D~\cite{mask3d} proposes a masked auto-encoder based self-supervised learning for ViT backbones to learn masked 3D priors for 2D image understanding without the reconstruction of RGB masked patches. In contrast, we formulate a self-supervised pre-training that can operate on both single-view images and videos and leverages masked 3D prior for several downstream tasks. Our approach also enhances the cross-modal learning and 2D-3D correspondence with contrastive learning and matching loss during pre-training which consequently boost the transfer learning performance.

\begin{figure*}
    \centering
    \includegraphics[width=1.0\textwidth]{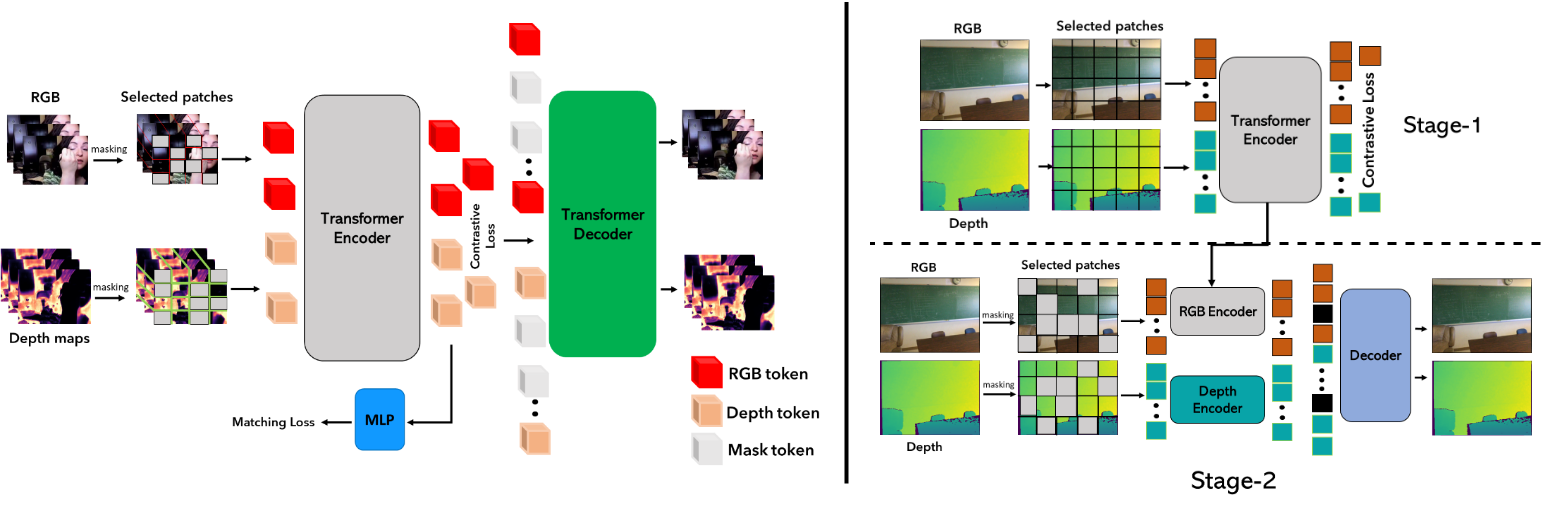}
    \caption{Overview of M$^{3}$3D pre-training. We introduce two self-supervised learning frameworks; Masked Image Modeling and contrastive learning. For MIM pre-training, we first mask out patches from RGB and depth input and then linearly project the unmasked patches to tokens with a fixed dimension. The tokens are then encoded using a Transformer encoder. The decoder takes the latent representations from the encoder and reconstruct the mask-out patches. Finally, we apply contrastive loss and matching loss on the encoded representations to enable cross-modal correspondence. \textbf{(Left)} M$^{3}$3D pre-training setup for video understanding. \textbf{(Right)} M$^{3}$3D two-stage pre-training setup for image understanding.}
    \label{fig:approach}
\end{figure*}

\section{Method}
\label{sec:approach}
We propose M$^{3}$3D to embed 3D priors, capture modality complementary feature and learn cross-modal representation by self-supervised pre-training using a single-view RGB-D image or a video clip. To learn such priors and representations, we formulate the problem from Masked autoencoder perspective and cross-modal training strategies. 

Given the RGB-D input, we first the masked the modalities, and pass them through the modality-specific or modality-agnostic encoder. The latent representations from the encoder(s) are then used to reconstruct the masked RGB and depth input. To learn the cross-modal representations, we use contrastive learning and matching loss which are the two most common cross-modal training strategies. The objective of depth reconstruction task is to embed geometric awareness into the encoder while cross-modal training enhances cross-modal representation learning. After pre-training, the encoder can be fine-tuned for downstream tasks such video action recognition, 2D semantic segmentation, depth estimation etc. An overview of our approach is show in Figure~\ref{fig:approach}. Below, we first explain the masking image modeling strategy from multi-modal input perspective. Then, we explain two types of cross-modal training strategies: contrastive, and matching loss. Finally, we describe the overall pre-training pipeline.

\subsection{Multi-modal Masked Autoencoder}
Our approach is not limited to the images and can be extended to the videos as well. Please see section~\ref{sec:exps} for performance on the video downstream tasks. For ease of notations, we describe our approach with a single-view RGB-D frame as an input. Given a RGB frame \textsc{I$_{R}$} of size \textsc{3 x H x W} and its corresponding depth map \textsc{I$_{D}$} of size \textsc{1 x H x W}, we first divide them into \textsc{P} x \textsc{P} patches from which we randomly keep a percentage of patches by masking out others for both RGB and depth input. \textsc{P} x \textsc{P} patches are projected to correct dimension using a different linear projection layer for each modality. Please refer to supplementary material on masking strategies. We use \texttt{Conv2d} (\texttt{Conv3d} in case of videos) as a linear projection layer. We then add positional embeddings and pass the tokens correspond to the unmasked patches to the modality-specific encoders or shared encoder between the modalities to get the latent representations. We use ViT-B as an encoder. To reconstruct the masked-out tokens, we use a lightweight shared decoder based on transformers with two prediction head; one for each modality. The input to the decoder is the latent representations of the unmasked patches and the set of mask tokens which are actually placeholders for the decoder to reconstruct the masked patches at the original image resolution. Following~\cite{multimae}, we add sine-cosine positional embeddings before passing them to the decoder. From the reconstruction output, we compute the losses only on the masked tokens.


\paragraph{Reconstruction Losses.} Our model has two reconstruction loss; one for each modality. Following~\cite{mae}, we first normalize the output patches as well as target patches, and then used compute the MSE loss between the ground truth and the predicted pixels. 

\begin{equation}
\mathcal{L}_{rgb} = \frac{1}{\omega} \sum_{p \in \omega} \vert \vert  \textsc{I$_{R}$} (p) - \hat{\textsc{I$_{R}$}}(p) \vert \vert_{2}
\label{eq:rgb loss}
\end{equation}
where $p$ is the token index, $\omega$ is the set of masked tokens. $\hat{\textsc{I$_{R}$}}$ corresponds to the reconstruction or the prediction of the model.

Similarly, for depth reconstruction task, we follow MultiMAE~\cite{multimae} to use L1 loss for image based pre-training. For video based pre-training, we use MSE Loss.
\begin{equation}
\mathcal{L}_{depth} = \frac{1}{\lambda} \sum_{m \in \lambda} \vert \vert  \textsc{I$_{D}$} (m) - \hat{\textsc{I$_{D}$}}(m) \vert \vert
\label{eq:depth loss}
\end{equation}
where $m$ is the token index, $\lambda$ is the set of masked tokens. $\hat{\textsc{I$_{D}$}}$ corresponds to the reconstruction or the prediction of the model.

\subsection{Cross-Modal Representation Learning}
Besides pre-training our model using masked image modeling, we also propose to use two cross-modal training strategies i.e. RGB-Depth contrastive learning and RGB-Depth matching. The objective of these strategies is to enhance the cross-modal representation learning.

\paragraph{Contrastive Loss.} Inspired by the success of contrastive learning~\cite{miech2020endtoend,clip_}, we present our cross-modal contrastive pre-training. For each RGB-Depth image pair, we use the latent representations from the transformer encoder. Specifically, we use InfoNCE loss~\cite{oord2019representation} with temperature $\tau$ to pre-train the model. The goal of RGB-D contrastive learning is to align the RGB patch and the corresponding depth patch by pulling them closer and repelling unpaired ones apart by minimizing the similarity with the other patches in the corresponding map. Indeed, one can use instance-level contrastive learning, but we hypothesize it will fail to capture high-level information as it is relatively an easy pre-text task.

\begin{equation}
\mathcal{L}_\mathrm{c} = - \frac{1}{N} \sum_{i=1}^N {\rm log}  \left[ \frac{ {\rm exp} (s_{i,i}/\tau)}{\sum_{k \neq i} {\rm exp} (s_{i,k}/\tau) + {\rm exp} (s_{i,i}/\tau)} \right]
\label{eq:contrastive loss}
\end{equation}
where $s_{i,j} = \|X^{rgb}_i\|^T\|X^d_j\|$ and $\tau$ is the temperature.  $\|X^{rgb}_i\|$ and $|X^d_j\|$ corresponds to the encoder features for RGB and depth respectively for a patch $i$.

\paragraph{Matching Loss.} Inspired by video-text matching, we also propose to use RGB-Depth matching to further enhance the cross-modal correspondence. It predicts whether the RGB-Depth pair is matched (positive) or not (negative). Specifically, we reuse the features from the encoder, and then passed them to the linear layer followed by a softmax classifier to solve two-class classification problem.

\begin{equation}
\mathcal{L}_\mathrm{m} = -\frac{1}{M} \sum_{i}^{M} \sum_{t=0}^{1} y_{it}^{\mathrm{m}} log(q_{t}^{\mathrm{m}}(X^{rgb}_i, X^d_i))
\label{eq:matching loss}
\end{equation}
where $y_{it}^{\mathrm{m}}$ is sign function which outputs 1 if value of t is 1 else 0. $q_{t}^{\mathrm{m}}(X^{rgb}_i, X^{d}_i)$ denotes the probability score of $t$ from the softmax. $M$ is the total number of RGB-Depth pairs in the batch. We specifically used this loss for video based pre-training.

\subsection{Pre-training Setup} The final pre-training loss of M$^{3}$3D for video based understanding is, 

\begin{equation}
\mathcal{L}_{video} = \alpha \mathcal{L}_{rgb} + \beta \mathcal{L}_{depth} + \gamma \mathcal{L}_\mathrm{c} + \eta \mathcal{L}_\mathrm{m}
\label{eq:total loss}
\end{equation}

where $\alpha$, $\beta$, $\gamma$, $\eta$ are the hyper-parameters tuned during pre-training. 
However, for image based understanding tasks, we pre-train our model in two-stages as shown in Figure~\ref{fig:approach}. In the first stage, we pre-train the encoder using contrastive loss for few epochs and then in the second stage, we initialize the weights of the encoder with stage-1 weights, and then pre-train the encoder and decoder using masked image modeling.

\begin{align}
    \mathcal{L}_{stage1} =  \mathcal{L}_\mathrm{c} \\ 
    \mathcal{L}_{stage2} = \alpha \mathcal{L}_{rgb} + \beta \mathcal{L}_{depth}
\label{eq:total loss image}
\end{align}
\section{Experiments}
\label{sec:exps}
M$^{3}$3D aims to learn masked 3D priors to embed to ViT backbones and enhance the fine-tuning performance with further cross-modal learning using contrastive learning and matching loss. In this section, we first briefly describe the experimental setup for pre-training stage. Then, we provide a transfer study to measure the effectiveness of our approach across various downstream tasks. Moreover, we show the data-efficient nature of our approach by evaluating it under low-data regime setting. Finally, we present some ablation studies to study the effectiveness of each component in our approach. 

\subsection{Experimental Setup}

\paragraph{Pre-train:} We use vision-transformer base (ViT-B) as a encoder for all the experiments. For video activity recognition, we pre-train the model on UCF-101~\cite{ucf101} dataset, one of the benchmark datasets for video action recognition. We use monodepth2~\cite{monodepth2} to extract the depths from the videos as original dataset contains RGB videos only. Please refer to~\cite{monodepth2} for more details on the approach. We use a lightweight decoder which consists of 4 blocks following VideoMAE~\cite{videomae}. We initialize the model weights with network weights trained on Kinetics-400~\cite{kinetics}. To maintain the self-supervised paradigm, we initialize the model with weights obtained by self-supervised Kinetics-400 pre-training~\cite{videomae}. We also show the efficacy of our approach by pre-training it from scratch. For surgical video action detection, we pre-train the model on OR-AR~\cite{OR-AR} dataset which consists of 820 full videos each having 9 temporal workflow phase labels. This dataset is collected using ToF cameras placed in two operating rooms in a single hospital. For 2D image understanding task, we pre-train the model on the ScanNet~\cite{scannet} dataset. ScanNet contains 2.5M RGB-D frames from 1513 video sequences. We regularly sample every 25$^{th}$ frame without any filtering during pre-training. We initialize the encoder weights with the self-supervised ImageNet pre-training~\cite{mae}. Please refer to the supplementary material for details on datasets, pre-training and hyper-parameters.

\subsection{Results}
\begin{table*}[h!]
\centering
\resizebox{0.80\textwidth}{!}{%
\begin{tabular}{c|c|c|c|c|c|c}
\hline

\hline

\hline
Methods  & Masking ratio  & Backbone &  Pre-train Epochs & Dataset & Frames & Top-1 \\
\hline

\hline

\hline
Scratch  & -                                                                       & ViT-B    & & UCF-101   & 16     & 51.4  \\ \hline

OPN~\cite{OPN}  & -                                                                       & VGG   & &UCF-101   & N/A     & 59.6  \\ \hline

VCOP~\cite{COP}  & -                                                                       & R(2+1)D  &   & UCF-101   & N/A     & 72.4  \\ \hline

CoCLR~\cite{coclr}  & -                                                                       & S3D-G  &  & UCF-101   & 32     & 81.4  \\ \hline

Vi$^{2}$CLR~\cite{Vi2clr}  & -                                                                       & S3D  &  & UCF-101   & 32     & 82.8 \\ \hline

CoCLR~\cite{coclr}  & -                                                                       & S3D-G   & & Kinetics-400   & 32     & 87.9  \\ \hline

Vi$^{2}$CLR~\cite{Vi2clr}  & -                                                                       & S3D     & & Kinetics-400   & 32     & 89.1 \\ \hline

MoCov3~\cite{Mocov3}   & -                                                                       & ViT-B    & & UCF-101   & 16     & 81.7  \\ \hline
VideoMAE~\cite{videomae} &                                                                     & ViT-B    &- & Kinetics-400   & 16     & 96.0$^{\dag}$ \\ \hline

\textbf{M$^{3}$3D}  & 0.90 & ViT-B    & 100 & Kinetics-400 + UCF-101   & 16     & \textbf{96.3}  \\
\hline

\hline

\hline
VideoMAE~\cite{videomae} & 0.75                                                                    & ViT-B    & 800 & UCF-101   & 16     & 90.1 \\ \hline
VideoMAE~\cite{videomae} & 0.90                                                                    & ViT-B    & 3200 & UCF-101   & 16     & 90.7 \\ \hline
VideoMAE~\cite{videomae} & 0.75                                                                    & ViT-B    & 3200 & UCF-101   & 16     & \textbf{91.2} \\ \hline
\textbf{M$^{3}$3D}  & 0.90 & ViT-B    & 800 & UCF-101   & 16     & \textbf{91.1}  \\ \hline
\end{tabular}
}%
\caption {\label{tab:ucf101} \textbf{Video action recognition accuracy on UCF-101~\cite{ucf101}}. $^{\dag}$ means we re-run the code using the original repository provided by VideoMAE.}
\end{table*}

\subsubsection{Fine-tuning for video action recognition}
We fine-tune M$^{3}$3D based ViT-B encoder on UCF-101. Table~\ref{tab:ucf101} shows the top-1 accuracy comparison with the recent state-of-the-art video based pre-training approaches. It can be seen from the table that our approach achieves 96.3\% top-1 accuracy outperforming all the baselines including VideoMAE~\cite{videomae} which shows the effectiveness of leveraging RGB-D data to embed 3D priors for video understanding. Moreover, we also report the results when we pre-train the model from scratch. Although, it achieves the same performance as VideoMAE, M$^{3}$3D is computationally less expensive than VideoMAE as it requires less epochs and higher masking ratio during pre-training. For fair comparison, we also report the performance of VideoMAE when it is pre-trained for less epochs with higher masking ratio (e.g. 0.90). It is clearly seen from the table that our approach outperforms VideoMAE under the same pre-training epochs and masking ratio. To the best of our knowledge, M$^{3}$3D provides the first ViT-B backbone that is pre-trained from scratch under self-supervision using RGB-D for video action recognition. It is worth mentioning that the depth maps are extracted from an off-the-shelf model and we hypothesize that the performance will further enhance with an improved depth estimation model or readily available depth data.

\begin{table*}[h!]
\centering
\resizebox{0.80\textwidth}{!}{%
\begin{tabular}{c|c|c|c|c|c|c|c}
\hline

\hline

\hline
Methods  & Masking ratio  & Backbone & Pre-train    & 5\%  & 10\%  & 20\% & 100\% \\ 
\hline

\hline

\hline
MaskFeat~\cite{maskfeat}      & 0.90                                                                  & MViT-S    & Kinetics-400 &62.35  & 78.88 &-  &- \\ \hline
MAE-Random~\cite{mae_vid}       & 0.90                                                                  & ViT-B    & Kinetics-400 & 64.66 & 81.48 &84.93 & 94.8\\ \hline
MAE-Random~\cite{mae_vid}       & 0.90                                                                  & ViT-B    & OR-AR & 66.58
 & 81.87 & 84.97& \textbf{96.3}\\ \hline
MAE-Frame~\cite{mae_vid}       & 0.90                                                                  &  ViT-B        & OR-AR    & 63.44 & 78.89 &81.45& -\\ \hline
VideoMAE~\cite{videomae} & 0.90                                                                    & ViT-B    & OR-AR   & 65.57 & 81.74& 83.89 &94.9\\ \hline
VideoMAE~\cite{videomae} & 0.90                                                                    & ViT-B    & Kinetics-400 + OR-AR   & 70.93 & 83.73 & 86.33 & 95.9\\ \hline 
\textbf{M$^{3}$3D}  & 0.90 & ViT-B    & Kinetics-400 + OR-AR    & \textbf{80.90} &\textbf{85.31} &\textbf{89.88} & 96.1 \\

\hline

\hline

\hline

\end{tabular}
}%
\caption {\label{tab:OR-AR} \textbf{Comparison of M$^{3}$3D with the other state-of the art methods under different data-regime setting on the OR-AR dataset~\cite{OR-AR}}.}
\end{table*}
\subsubsection{Fine-tuning for surgical video action detection}
We fine-tune M$^{3}$3D based ViT-B encoder on OR-AR, a benchmark dataset for surgical video action detection. Table~\ref{tab:OR-AR} shows the results compared to the recent state-of-the-art video based masked autoencoder pre-training approaches. Following common practice~\cite{OR-AR,aidean}, we report the mean average precision (mAP). Our approach consistently outperforms the existing video based MAE models on low-data regime. More specifically, M$^{3}$3D achieves 80.90 mAP when fine-tuned using 5\% labeled data which also shows that it is a more data-efficient learner than VideoMAE. Moreover, we report the results of VideoMAE when it is initialized using Kinetics-400 pre-trained weights. The weights can be found from VideoMAE's github repository. Similar to video action recognition, it is the first multi-modal based transformer for surgical action detection which shows the effectiveness of the M$^{3}$3D on datasets in the medical domain. Finally, we report the results of Spatio-temporal MAE~\cite{mae_vid} with different masking strategies such as random, frame, etc. and they are termed as MAE-Random, MAE-Frame in the table.

\begin{table*}[h!]
\centering
\resizebox{0.80\textwidth}{!}{%
\begin{tabular}{c|c|c|c|c|c}
\hline

\hline

\hline
Methods  & Reconstruction task  & Backbone &  Pre-train  & Fine-tune Modality  & mIoU \\
\hline

\hline

\hline
Scratch  & -                                                                       & ViT-B    & \textcolor{gray}{None} & RGB     & 32.6  \\ \hline

MultiMAE*~\cite{multimae}  & RGB + Depth                                                                       & ViT-B    & ImageNet+ScanNet&RGB       & 64.0  \\ \hline

Pri3D~\cite{pri3d_}  &                                                                        & ViT-B   & ImageNet+ScanNet  & RGB      & 59.3  \\ \hline

Pri3D~\cite{pri3d_}  & -                                                                       & ResNet-50   &  ImageNet+ScanNet & RGB        & 60.2  \\ \hline

DINO~\cite{dino}  & -                                                                       & ViT-B   & ImageNet+ScanNet  & RGB        & 58.1 \\ \hline

MAE~\cite{mae}  & RGB                                                                      & ViT-B    &  ImageNet& RGB        & 64.8  \\ \hline

MAE~\cite{mae}  & RGB                                                                       & ViT-B      & ImageNet+ScanNet  & RGB       & 64.5 \\ \hline

Mask3D~\cite{mask3d}   & Depth                                                                       & ViT-B    & ImageNet+ScanNet       & RGB        & 66.0  \\ \hline
Mask3D~\cite{mask3d} &  RGB + Depth                                                                   & ViT-B    & ImageNet+ScanNet  & RGB & 65.5 \\ \hline

\textbf{M$^{3}$3D}  & RGB + Depth & ViT-B    & ImageNet+ScanNet   & RGB        & \textbf{67.3}  \\
\hline

\hline

\hline
\rowcolor{lightgray}
MultiMAE~\cite{multimae} & RGB + Depth + Segmentation & ViT-B & ImageNet  & RGB   & 66.4 \\ \hline
\end{tabular}
}%
\caption {\label{tab:scannet} \textbf{ScanNet Semantic Segmentation}. M$^{3}$3D outperforms Mask3D and other state-of-the-art approaches that leverage RGB-D data during pre-training.}
\end{table*}

\begin{table*}[h!]
\centering

\resizebox{0.80\textwidth}{!}{%
\begin{tabular}{c|c|c|c|c|c}
\hline

\hline

\hline
Methods  & Reconstruction task  & Backbone &  Pre-train  & Fine-tune Modality  & mIoU \\
\hline

\hline

\hline

MultiMAE*~\cite{multimae}  & RGB + Depth                                                                       & ViT-B    & ImageNet+ScanNet& RGB       & 49.1  \\ \hline

MAE~\cite{mae}  & RGB                                                                      & ViT-B    &  ImageNet& RGB        & 46.9  \\ \hline

MAE~\cite{mae}  & RGB                                                                       & ViT-B      & ImageNet+ScanNet  & RGB       & 48.3 \\ \hline

Mask3D~\cite{mask3d}   & Depth                                                                       & ViT-B    & ImageNet+ScanNet       & RGB        & 50.5  \\ \hline

\textbf{M$^{3}$3D}  & RGB + Depth & ViT-B    & ImageNet+ScanNet   & RGB        & \textbf{51.2}  \\
\hline

\hline

\hline
\rowcolor{lightgray}
MultiMAE~\cite{multimae} & RGB + Depth + Segmentation & ViT-B & ImageNet  & RGB       & \textbf{51.5}$^{\dag}$ \\ \hline
\end{tabular}
}%
\caption {\label{tab:nyu2} \textbf{NYUv2 Semantic Segmentation}. M$^{3}$3D outperforms state-of-the-art approaches that leverage RGB-D data during pre-training. It demonstrate the effectiveness in transferring to out-domain dataset. $^{\dag}$ means we re-run the code using the original repository provided by MultiMAE. }
\end{table*}
\subsubsection{Fine-tuning for 2D semantic segmentation}
In this section, we show that our pre-training approach is capable of transferring the learned representations to the 2D image understanding tasks by performing the 2D semantic segmentation task. Following MultiMAE~\cite{multimae}, we use segmentation head based on ConvNext architecture~\cite{convnext} on top of the ViT-B encoder. We use mean Intersection over Union (mIoU) as an evaluation metric. For fine-tuning, we sample every 100$^{th}$ frame, resulting in 20,000 training images and 5000 validation images which is common protocol of the ScanNet benchmark~\cite{scannet}. Table~\ref{tab:scannet} shows the results compared to the recent state-of-the-art pre-training approaches for ScanNet semantic segmentation. It can be observed from the table that M$^{3}$3D significantly outperforms Mask3D~\cite{mask3d} (\textbf{+1.3} mIoU), a state-of-the-art approach. More notably, our approach improves over MAE~\cite{mae} which is also pre-trained with ImageNet and ScanNet (+2.8 mIoU). Moreover, compared to Pri3D~\cite{pri3d_}, a 3D-based pre-training method, our approach outperforms by a significant margin of \textbf{7.1} mIoU which shows that multi-view based 3D pre-training does not effectively embed 3D priors to ViT backbones, rather degrades the performance compared to ResNet backbone. Furthermore, we also report the results of MultiMAE~\cite{multimae} when it is pre-trained on ScanNet and it is denoted as MultiMAE* in the table. For fair comparison, we only use RGB and depth as the reconstruction task. We observe that M$^{3}$3D outperforms MultiMAE* by a margin of \textbf{3.3} mIoU as well as original MultiMAE . Finally, we demonstrate the generalizability of M$^{3}$3D across datasets by fine-tuning the pre-trained model on NYUv2~\cite{nyuv2} following the same setup. Table~\ref{tab:nyu2} shows the results compared to the recent state-of-the-art pre-training approaches for NYUv2 semantic segmentation. We draw the same observation that our pre-trained approach outperforms the competing baselines showing how well it transfers across datasets.


\begin{table*}[h!]
\centering

\resizebox{0.80\textwidth}{!}{%
\begin{tabular}{c|c|c|c|c|c}
\hline

\hline

\hline
Methods  & Reconstruction task  & Backbone &  Pre-train  & Fine-tune Modality  & $\delta_{1}$ \\
\hline

\hline

\hline

MultiMAE~\cite{multimae}  & RGB + Depth                                                                       & ViT-B    & ImageNet+ScanNet& RGB       & 85.3  \\ \hline

MAE~\cite{mae}  & RGB                                                                      & ViT-B    &  ImageNet& RGB        & 85.1  \\ \hline

Mask3D~\cite{mask3d}   & Depth                                                                       & ViT-B    & ImageNet+ScanNet       & RGB        & 85.4  \\ \hline

CroCo~\cite{croco}  & RGB + Depth                                                                       & ViT-B    & Habitat& RGB       & 85.6  \\ \hline

MultiMAE*~\cite{multimae}  & RGB + Depth + Segmentation                                                                      & ViT-B    & ImageNet & RGB       & 83.0  \\ \hline

\textbf{M$^{3}$3D}  & RGB + Depth & ViT-B    & ImageNet+ScanNet   & RGB        & \textbf{86.7}  \\
\hline

\hline

\hline
\rowcolor{lightgray}
MultiMAE~\cite{multimae} & RGB + Depth + Segmentation & ViT-B & ImageNet  & RGB       & \textbf{86.4} \\ \hline
\end{tabular}
}%
\caption {\label{tab:nyu2_depth} \textbf{NYUv2 Depth Estimation}. M$^{3}$3D outperforms Mask3D and other state-of-the-art approaches that leverage RGB-D data during pre-training. It demonstrate the effectiveness in transferring to dense prediction task for out-domain dataset.* means the number are reported from CroCo~\cite{croco} paper.}
\end{table*}
\subsubsection{Fine-tuning for depth estimation}
In this section, we study how M$^{3}$3D transfers the representation to the dense regression tasks. We use NYUv2 for the depth estimation and report $\delta_{1}$ on the NYUv2 test set. $\delta_{1}$ is the percentage of pixels that have an error ratio ($\max\{ \frac{\hat{y}_{p}}{y_{p}}, \frac{y_{p}}{\hat{y}_{p}} \}$) below 1.25~\cite{delta1_eval}. Following MultiMAE, we use DPT~\cite{DPT} as dense prediction head on the top of the ViT encoder. Table~\ref{tab:nyu2_depth} shows that M$^{3}$3D outperforms recent state-of-the-art approaches including MultiMAE, Mask3D. Notably, it outperforms CroCo~\cite{croco} which is a cross view completion pre-training strategy based on MIM specifically designed for 3D vision tasks with 86.7\% vs 85.6\%. Although our approach is marginally outperforming original MultiMAE, but we want to reiterate that it is pre-trained with three different tasks. Moreover, when multiMAE is pre-trained on ScanNet with RGB and depth only, the performance drops to 85.3\%. 

\begin{figure}
    \centering
    \includegraphics[width=0.45\textwidth]{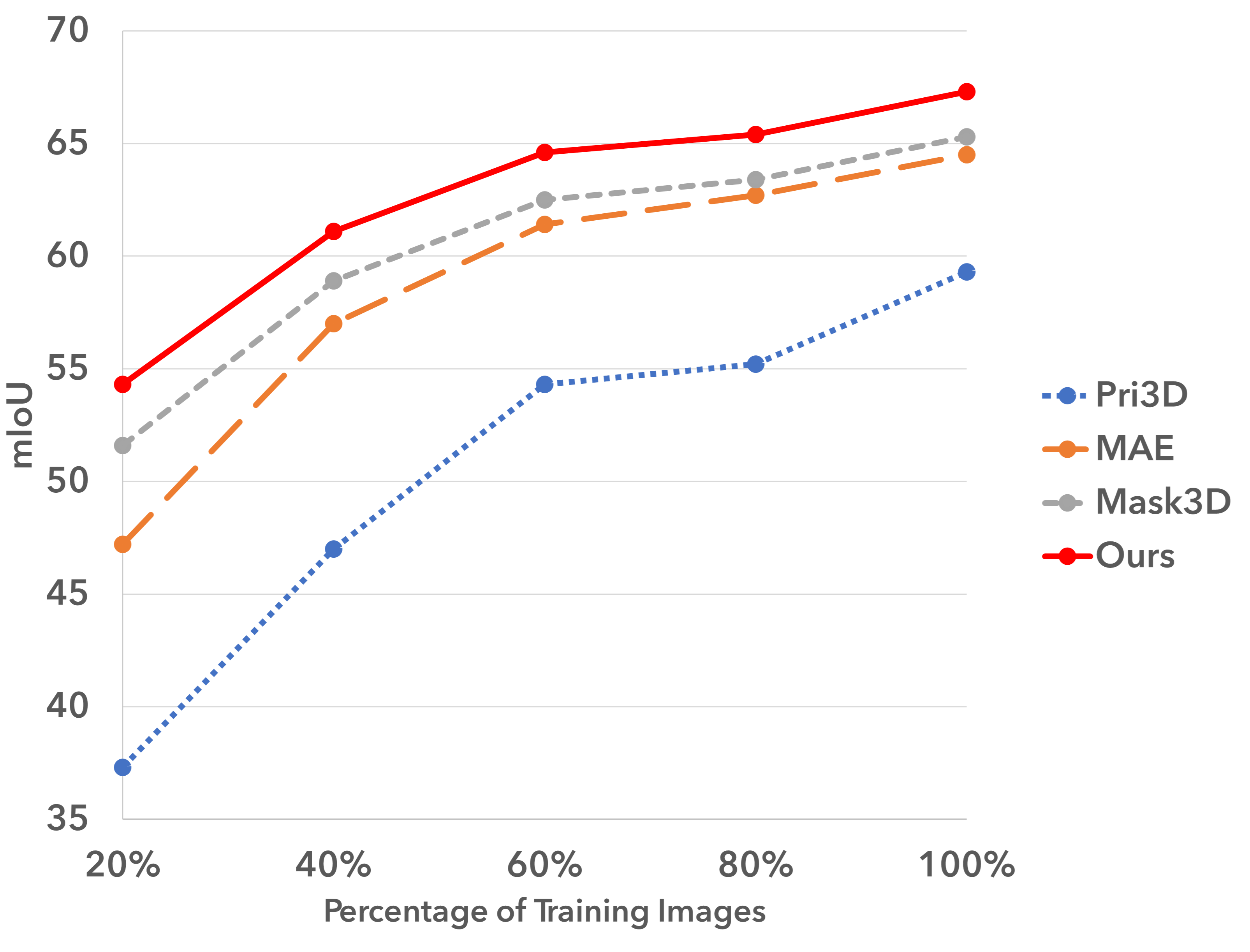}
    \caption{M$^{3}$3D is a data-efficient learner. We compare to the recent state-of-the-art pre-training approaches on ScanNet 2D semantic segmentation under limited labeled data scenarios. Notably, our approach improves +2.7\% mIoU over Mask3D~\cite{mask3d} at 20\% training data. }
    \label{fig:data_efficient}
\end{figure}
\subsection{Data-Efficient Learner}
We fine-tune M$^{3}$3D based ViT-B encoder on ScanNet for 2D semantic segmentation mainly under low-data regime setting to study how data-efficient learner our approach is. Figure~\ref{fig:data_efficient} shows that our approach consistently outperforms the competing approaches by a considerable margin across different percentage of available labeled training data. It is worth mentioning that M$^{3}$3D recovers more than 80\% performance of the full labeled training set performance when fine-tune with only 20\% of training data.


\subsection{Ablation studies}
In this section, we perform in-depth ablation studies on ScanNet 2D semantic segmentation. 
\vspace{-10pt}
\paragraph{Effect of masking ratio during pre-training.}
\begin{table}[h!]
\centering

\begin{tabular}{|c|c|c|}
\hline

\hline

\hline
RGB ratio  & Depth ratio  & mIoU \\
\hline

\hline

\hline

20.0\% & 20.0\% & 65.5\\ \hline

20.0\% & 50.0\% & 65.4\\ \hline

20.0\% & 80.0\% & 65.4\\ \hline

50.0\% & 20.0\% & 66.0\\ \hline

50.0\% & 50.0\% & 65.3\\ \hline

80.0\% & 20.0\% & 66.1\\ \hline

80.0\% & 80.0\% & \textbf{67.3} \\ \hline

\end{tabular}
\caption {\label{tab:ablation_masking} We study the effect of different masking ratio for RGB and depth input on ScanNet 2D semantic segmentation where each ratio indicates the percentage of masked patches.}
\end{table}
We study the influence of masking ratio and report the results in Table~\ref{tab:ablation_masking} on ScanNet semantic segmentation. It is clearly shown from the table that by masking more patches in RGB and depth modality, M$^{3}$3D achieves the best performance.

\paragraph{Without ImageNet Initialization.}
\begin{table}[h!]
\centering

\begin{tabular}{|c|c|c|}
\hline

\hline

\hline
Method  & Backbone  & mIoU \\
\hline

\hline

\hline

Train from Scratch & ViT-B & 32.6\\ \hline

Mask3D~\cite{mask3d} & ViT-B & 41.3\\ \hline

M$^{3}$3D & ViT-B & \textbf{42.5} \\ \hline

\end{tabular}
\caption {\label{tab:ablation_init} Results on ScanNet 2D semantic segmentation without ImageNet initialization during pre-training.}
\end{table}
We observe a performance drop when the model is not initialized using ImageNet pre-trained weights in the pre-training stage as shown in Table~\ref{tab:ablation_init}. Because ScanNet has a relatively small amount of indoor data, so it's harder to pre-train the ViT backbones from scratch. Since, ImageNet weights are readily available, so we initialize our models with it following~\cite{mask3d,pri3d_}.

\paragraph{Effect of each loss function.}
\begin{table}[h!]
\centering

\begin{tabular}{|c|c|c|}
\hline

\hline

\hline
Loss  & Backbone  & mIoU \\
\hline

\hline

\hline

w/out contrastive loss & ViT-B & 65.5\\ \hline

w/out RGB task & ViT-B & 66.8 \\ \hline

M$^{3}$3D & ViT-B & \textbf{67.3} \\ \hline

\end{tabular}
\caption {\label{tab:ablation_loss} Results on on ScanNet 2D semantic segmentation by pre-training without different loss functions.}
\end{table}
We also study the effect of loss functions during pre-training and report the performance on ScanNet 2D semantic segmentation. We compare the performance of model without contrastive loss and without RGB reconstruction task. The results are reported in Table~\ref{tab:ablation_loss}. From the results, we observe that contrastive loss overall improves the mIoU compared to Mask3D which suggests that cross-modal learning is an important component besides Masked Image Modeling. 

\begin{figure}
    \centering
    \includegraphics[width=0.47\textwidth]{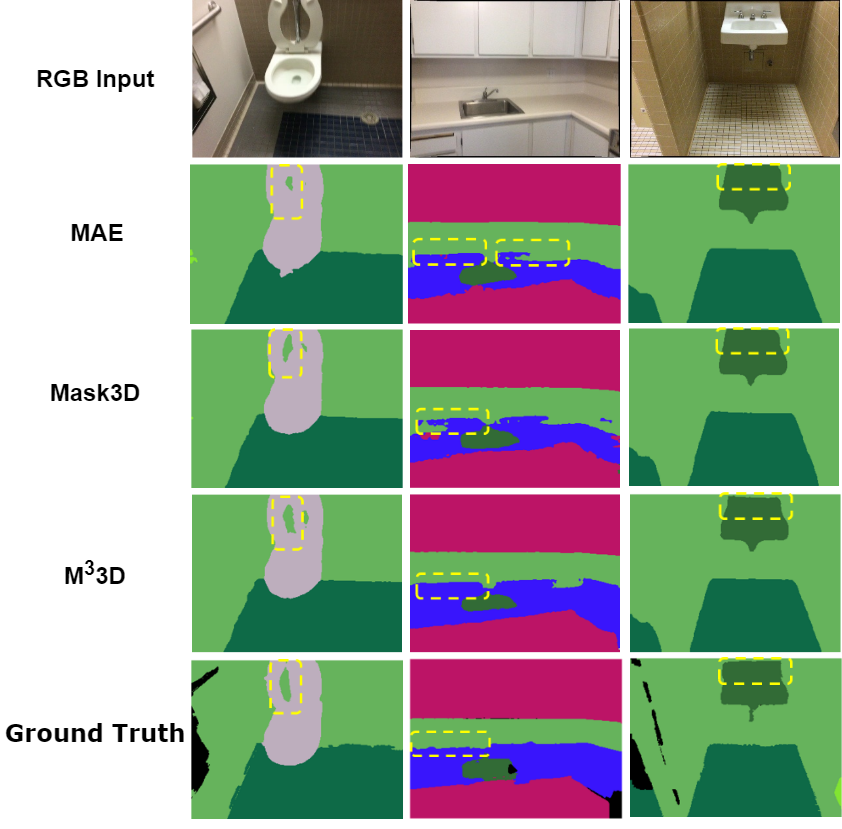}
    \caption{\textbf{Qualitative Results on ScanNet.} We visualize the predictions of various approaches on 2D Semantic Segmentation task. }
    \label{fig:visualize}
    \vspace{-10pt}
\end{figure}
\section{Conclusion}
In this paper, we present M$^{3}$3D, a new self-supervised pre-training technique with two main functions: (1) learning to embed geometric priors into 2D representations using Masked Image Modeling, (2) learning cross-modal representations in RGB-D data using contrastive learning. M$^{3}$3D is a general pre-training method applicable to a variety of image/video understanding tasks, doesn't require camera registration between multi-view input as found in recent self-supervised approaches such as Pri3D, and works well in low-data regime. Our extensive experiments on downstream tasks such as video action recognition, video action detection, 2D semantic segmentation and depth estimation show the superiority of  M$^{3}$3D compared to current state-of-the-art 2D representation learning approaches. Future work includes extending the approach to applications with datasets that have more than two modalities.

{\small
\bibliographystyle{ieee_fullname}
\bibliography{references}
}

\clearpage
\appendix
\section*{Appendices}
\setcounter{section}{0}
\def\thesection{\Alph{section}}

\section{Datasets}

\paragraph{Video Understanding.} For video action recognition, we pre-train and evaluate the model on UCF-101~\cite{ucf101} which contains 9.5k/3.5k train/val videos. For surgical video action detection, we pre-train and evaluate the model on OR-AR~\cite{OR-AR} which consists of 820 long videos captured in surgical operating rooms. All videos have 9 temporal workflow phase labels.
\vspace{-10pt}
\paragraph{Image Understanding.} For image understanding tasks, we pre-train the model on ScanNet~\cite{scannet} which contains 2.5M RGB-D frames from 1513 video sequences. For evaluation, we use ScanNet and NYUv2~\cite{nyuv2}. NYUv2 contains 1449 densely labeled images from indoor scenes captured with Microsoft Kinect RGB-D camera. We use the official split of 795 images in training set and 654 images in test set.

\section{Additional Implementation Details} \label{implementations}

\subsection{Network Architecture}

\paragraph{Video Understanding.} We follow the network architecture presented in VideoMAE~\cite{videomae} for video based pre-training. The encoder part of the network is vision-transformer base (ViT-B) while the decoder consists of 4 blocks with six multi-head attentions in each. The width of the decoder is set to half of the encoder dimension i.e. 384-d. We use two fully connected layers (one for each modality) on top of the decoder for reconstruction. During fine-tuning, we remove the decoder and add the fully connected layer for prediction. Specifically for surgical video action detection, we follow the fine-tuning method in~\cite{OR-AR,aidean}. For evaluation, we fine-tune the model in two stages. In the first stage, we add a fully connected layer on top of the encoder for predicting clip-wise phases. In the second stage, we first extract the features from the encoder and then train a temporal model (Bi-GRU) for detecting phases in a full video.
\vspace{-10pt}
\paragraph{Image Understanding.} We follow MAE~\cite{mae} for the network design. Our modality-specific encoders are based on ViT-B while the decoder part consists of 8 blocks with 16 multi-head attentions in each block. The width is set to 512. Similarly, we use two fully-connected layers (one for each modality) on top of the decoder for reconstruction. For fine-tuning, we mostly follow MultiMAE~\cite{multimae} for task specific head. More specifically, we use segmentation head based on ConvNeXt architecture~\cite{convnext} and depth-estimation head based on DPT~\cite{DPT}.

\subsection{Pre-training and Fine-tuning Details}
For video understanding, we report the pre-training setting in Table~\ref{tab:pretrain} and the fine-tuning setting in Table~\ref{tab:fine-tune}. Moreover, we report the pre-training setting on ScanNet in Table~\ref{tab:pretrain_scannet} and the transfer setting for semantic segmentation and depth estimation task in Table~\ref{tab:fine-tune_segmentation} and Table~\ref{tab:fine-tune_depth} respectively.

\begin{table}[h]
    \centering
    
    \resizebox{0.45\textwidth}{!}{%
    \begin{tabular}{l|cc}
    \hline
    
    \hline
    
    \hline
    Configuration & OR-AR~\cite{OR-AR} & UCF-101~\cite{ucf101} \\
    \hline
    
    \hline
    
    \hline
    Optimizer & \multicolumn{2}{c}{AdamW}\\
    Optimizer betas & \multicolumn{2}{c}{\{0.9, 0.95\}}\\
    Base learning rate & 1e-4 & 1e-3 \\
    Weight decay & \multicolumn{2}{c}{5e-2}\\
    Learning rate schedule & \multicolumn{2}{c}{cosine decay} \\
    gradient clipping & 0.02 & None \\
    Warmup epochs & \multicolumn{2}{c}{40}\\
    Epochs & 1600 & 100 or 800 (from scratch) \\
    \hline
    
    \hline
    Flip augmentation & True & True \\
    Augmentation & \multicolumn{2}{c}{MultiScaleCrop}\\
    Num of Frames & \multicolumn{2}{c}{16}\\
    sampling rate & \multicolumn{2}{c}{4.0}\\
    \hline
    
    \hline
    $\alpha$ & 1.0 & 1.0\\
    $\beta$ & 0.5 & 0.1\\
    $\gamma$ & 0.2 & 0.01\\
    $\eta$ & 0.1 & 0.01\\
    \hline
    
    \hline
    
    \hline
    \end{tabular}
    }%
    \caption{Pre-training setting on OR-AR~\cite{OR-AR} and UCF-101~\cite{ucf101} datasets.}
    \label{tab:pretrain}
\end{table}

\begin{table}[h]
    \centering
    
    \resizebox{0.45\textwidth}{!}{%
    \begin{tabular}{l|cc}
    \hline
    
    \hline
    
    \hline
    Configuration & OR-AR~\cite{OR-AR} & UCF-101~\cite{ucf101}\\
    \hline
    
    \hline
    
    \hline
    Optimizer & \multicolumn{2}{c}{AdamW}\\
    Optimizer betas & \multicolumn{2}{c}{\{0.9, 0.95\}}\\
    Base learning rate & 6e-4 & 1e-3\\
    Weight decay & \multicolumn{2}{c}{5e-2}\\
    Learning rate schedule & \multicolumn{2}{c}{cosine decay} \\
    Warmup epochs & \multicolumn{2}{c}{5}\\
    Epochs & 75 & 100\\
    \hline
    
    \hline
    Flip augmentation & True & True \\
    Mixup & None & 0.8 \\
    CutMix & None & 1.0 \\
    drop path & 0.1 & 0.2 \\
    drop out & 0.0 & 0.5 \\
    Layer-wise lr decay & 0.65 & 0.70\\
    \hline
    
    \hline
    
    \hline
    Temporal Model learning rate & 1e-3 & None \\
    Temporal Model Epochs & 15 & None\\
    \hline
    
    \hline
    
    \hline
    \end{tabular}
    }%
    \caption{Fine-tune setting on OR-AR~\cite{OR-AR}, UCF-101~\cite{ucf101} datasets.}
    \label{tab:fine-tune}
\end{table}


\begin{table}[h]
    \centering
    
    \resizebox{0.50\textwidth}{!}{%
    \begin{tabular}{l|ccc}
    \hline
    
    \hline
    
    \hline
    Strategy and Ratio & OR-AR~\cite{OR-AR} & UCF-101~\cite{ucf101} & ScanNet~\cite{scannet}\\
    \hline
    
    \hline
    
    \hline
    RGB Masking strategy &  Tube & SurgMAE~\cite{surgmae} & Random \\
    RGB Masking ratio & 0.9 & 0.9 & 0.8 \\
    Depth Masking strategy & Tube & Random & Random \\
    Depth Masking ratio & 0.9 & 0.9 & 0.8\\

    \hline
    
    \hline
    
    \hline

    \end{tabular}
    }%
    \caption{Masking strategies during pre-training.}
    \label{tab:masking_strat}
\end{table}


\begin{table}[h]
    \centering
    
    \resizebox{0.45\textwidth}{!}{%
    \begin{tabular}{l|c}
    \hline
    
    \hline
    
    \hline
    Configuration & ScanNet~\cite{scannet} \\
    \hline
    
    \hline
    
    \hline
    Optimizer & AdamW\\
    Optimizer betas & {\{0.9, 0.95\}}\\
    Base learning rate & 1e-4 \\
    Weight decay & {5e-2}\\
    Learning rate schedule & {cosine decay} \\
    Stage-1 epochs & 20 \\
    Stage-2 epochs & 100 \\
    \hline
    
    \hline
    Augmentation & {Gaussian Blur, ColorJitter}\\
    $\alpha$ & 0.1 \\
    $\beta$ & 1.0 \\

    \hline
    
    \hline
    
    \hline
    \end{tabular}
    }%
    \caption{Pre-training setting on ScanNet~\cite{scannet}.}
    \label{tab:pretrain_scannet}
\end{table}

\begin{table}[h]
    \centering
    
    \resizebox{0.45\textwidth}{!}{%
    \begin{tabular}{l|cc}
    \hline
    
    \hline
    
    \hline
    Configuration & ScanNet~\cite{scannet} & NYUv2~\cite{nyuv2}\\
    \hline
    
    \hline
    
    \hline
    Optimizer & \multicolumn{2}{c}{AdamW}\\
    Optimizer betas & \multicolumn{2}{c}{\{0.9, 0.999\}}\\
    Base learning rate & \multicolumn{2}{c}{1e-4}\\
    Layer-wise lr decay & \multicolumn{2}{c}{0.75}\\
    Weight decay & \multicolumn{2}{c}{5e-2}\\
    Learning rate schedule & \multicolumn{2}{c}{cosine decay} \\
    Warmup epochs & \multicolumn{2}{c}1\\
    Warmup learning rate & \multicolumn{2}{c}{1e-6} \\
    Drop path & \multicolumn{2}{c}{0.1} \\
    Epochs & 50 & 200\\
    \hline
    
    \hline
    Input resolution & 240 x 320 & 640 x 640 \\
    Color jitter & \xmark & \cmark \\
    RandomGaussianBlur & \cmark  & \xmark \\
    RandomHorizontalFlip & \cmark  & \xmark \\    
    \hline
    
    \hline
    
    \hline
    \end{tabular}
    }%
    \caption{Fine-tune setting on ScanNet~\cite{scannet} and NYUv2~\cite{nyuv2} for 2D semantic segmentation.}
    \label{tab:fine-tune_segmentation}
\end{table}

\begin{table}[h]
    \centering
    
    \resizebox{0.45\textwidth}{!}{%
    \begin{tabular}{l|c}
    \hline
    
    \hline
    
    \hline
    Configuration & NYUv2~\cite{nyuv2} \\
    \hline
    
    \hline
    
    \hline
    Optimizer & {AdamW}\\
    Optimizer betas & {\{0.9, 0.999\}}\\
    Base learning rate & 1e-4\\
    Weight decay & 1e-4 \\
    Learning rate schedule & {cosine decay} \\
    Warmup epochs & 100\\
    Warmup learning rate & 1e-6 \\
    Epochs & 2000 \\
    Batch Size & 128 \\
    Layer-wise lr decay & 0.75\\
    \hline
    
    \hline
    Input resolution & 256 x 256 \\
    Augmentation & {RandomCrop, Color jitter} \\
    
    \hline
    
    \hline
    
    \hline
    \end{tabular}
    }%
    \caption{Fine-tune setting for NYUv2~\cite{nyuv2} depth estimation.}
    \label{tab:fine-tune_depth}
\end{table}

\subsection{Masking Strategy}
Table~\ref{tab:masking_strat} shows the different masking strategies for RGB-D modalities during pre-training.

\end{document}